\documentclass[conference,a4paper]{IEEEtran}
\usepackage[textwidth=16.5cm,left=20mm,
 top=20mm, bottom=20mm]{geometry}

\IEEEoverridecommandlockouts                              

\usepackage{multirow}
\usepackage{url}
\usepackage{float}
\usepackage[utf8]{vietnam}
\usepackage[vietnamese,english]{babel}
\usepackage{graphicx}
\usepackage{lipsum}
\title{\LARGE \bf
Deep Learning versus Traditional Classifiers on \\ Vietnamese Students' Feedback Corpus
}

\author{\textbf{Phu X. V. Nguyen$^{1}$, Tham T. T. Hong$^{2}$, Kiet Van Nguyen$^{3}$ , Ngan Luu-Thuy Nguyen$^{4}$} \\
University of Information Technology , Vietnam National University \\
Ho Chi Minh City \\
14520685@gm.uit.edu.vn$^{1}$, 14520824@gm.uit.edu.vn$^{2}$, kietnv@uit.edu.vn$^{3}$, ngannlt@uit.edu.vn$^{4}$
}

\begin{document}
\setlength\columnsep{0.25in}
\maketitle

\thispagestyle{empty}
\pagestyle{empty}

\begin{abstract}

Student's feedback is an important source of collecting students’ opinions to improve quality of training activities. Implementing sentiment analysis into student feedback data, we can determine sentiments polarities which express all problems in the institution since changes necessary will be applied to improve the quality of teaching and learning. This study focused on the machine learning and natural language processing techniques (Naive Bayes, Maximum Entropy, Long Short-Term Memory, Bi-Directional Long Short-Term Memory) on the Vietnamese Students' Feedback Corpus collected from a university. The final results were compared and evaluated to find the most effective model based on different evaluation criteria. The experimental results show that Bi-Directional Long Short-Term Memory algorithm outperformed than three other algorithms in term of the F1-score measurement with 92.0\% on the sentiment classification task and 89.6\% on the topic classification task. In addition, we developed a sentiment analysis application analyzing student feedback. The application will help the institution to recognize students' opinions about a problem and identify shortcomings that still exist. With the use of this application, the institution can propose an appropriate method to improve the quality of training activities in the future.

\end{abstract}

\section{INTRODUCTION}

In recent years, the quality of training activities in the institution has been noticed and gradually improved in order to extend the student's ability to acquire knowledge. The quality of training activities in the institution is not evaluated by the qualifications or knowledge of the lecturers but must be evaluated from the satisfaction of the students. To do this, it is necessary to know how students learn and think about the problems that exist in the institution. Feedback is often used for this purpose. Feedback may be categorized by: (1) Feedback from faculty to students, used for student self-assessment. (2) Feedback from students and teachers, allowing them to express their ideas in order to improve their teaching and improve their knowledge acquisition. This is one of the most important information to help the institution overcome the weaknesses and improve its training program. The collection of student feedback will reflect the shortcomings that arise in the training activities (curriculums, lecturers, facilities, others) and students' opinions (positive, negative, neutral). \\
 There are many ways to evaluate the quality of training activities based on student feedback. For example, in some teacher evaluation systems, the quantitative data were collected by closed-ended questions such multiple choices, while the qualitative data was collected by open-ended questions as comments and suggestions from students opinion in a textual form \cite{c1}, or it could be a score based on pre-determined criteria. Then the evaluation has been done manually, mainly based on people which is very laborious and time-consuming. The amount of feedback collected can be very large and complicated. Information is sometimes overlooked, which can be vital to evaluate the quality of training activities. Because of these significant reasons, it is very important to build a sentiment analysis application that automatically classifies student feedback. It helps institutions improve the quality of training activities, minimizing time and costs. \\
 Sentiment analysis, also called opinion mining, is the field of study that analyzes people’s opinions, sentiments, evaluations, appraisals, attitudes, and emotions towards entities such as products, services, organizations, individuals, issues, events, topics, and their attributes \cite{c2}.  Sentiment analysis of student feedback is a form of indirect assessment that analyzes text written by students—whether in formal course surveys or informal comments from online platforms to determine students’ interest in a class and to identify areas that could be improved through corrective actions\cite{c3}.  \\
Our study mainly focused on applying deep learning and traditional classifiers implemented on the student feedback corpus in Vietnamese. By conducting experiments and analyzing the results, we selected the most effective model for developing the student feedback analysis application. This application is based on deep learning and capable of analyzing student feedback and outputting results based on the student topic mentioned (curriculums, lecturers, facilities, others) and opinions of the students (positive, negative, neutral). Institutions can use these results to resolve existing shortcomings and improve the quality of training activities in the next semester. 
\\
In this study, we preform two tasks:
\begin{itemize}
\item Based on student feedback, analyze the sentiment polarity, determine which label that the feedback belongs to (positive, negative, neutral).\\
For example, given the feedback: "Giảng\_viên dạy rất hay và tâm\_huyết" meaning "Lecturers teach very well and enthusiastic" our task is to determine the corresponding sentiment label (positive).
\item Based on student feedback, analyze the topic polarity, determine which label that the feedback belongs to (curriculums, lecturers, facilities, others). \\
For example, given the feedback "Máy vi\_tính trong phòng thực\_hành đôi khi gặp trục\_trặc" meaning "The computers in the lab sometimes have problems" our task is to determine the corresponding topic label (facilities).
\end{itemize}

 The structure of the paper is presented as follows. In section 2, we will show an overview of the study of sentiment analysis and especially in the field of education. In section 3, we will present the detailed description of the corpus. In section 4, We will briefly introduce two traditional classifiers and two deep learning classifiers. Experimental settings will be presented in session 5. The results will be evaluated and analyzed in section 6. In the next section, the student feedback analysis application will be introduced. Finally, we will come to the conclusion and future works in section 7.
\section{PREVIOUS WORKS}
In recent years, the field of sentiment analysis has been growing and there are many published research papers especially in the field of education. 
In \cite{c4}, V. Dhanalakshmi et al., proposed to build a system which was applied machine learning and natural language processing techniques in a student feedback evaluation module from Middle East College, Oman. The data used is student feedback about 6 programs in the institution. There were 4 algorithms used to classify feedback which were Support Vector Machine, Naive Bayes, K-Nearest Neighbor and Neural Network. The result was evaluated on 10-fold cross validation method. The experimental results show that K-Nearest Neighbor performed the best in term of precision which was 100\% and Naive Bayes achieved 99.07\% and 99.11\% in term of precision and recall respectively. \\
Similarly \cite{c5}, A. Ortigosa et al,. proposed a new method for sentiment analysis on Facebook. By extracting information from user messages, they visualized emotion status of the user to determine the significant change in emotion. After that, the results were applied to an adaptive e-learning system to support personalized learning by considering user emotion and recommending suitable action for each case. In the study, they combined lexical-based and machine learning method in Tree-decision, Naive Bayes, and SVM and obtained 83.17\% in term of accuracy for the hybrid method (Tree decision and Lexical-based). \\
In \cite{c6}, they proposed a system for analyzing, processing and visualizing emotion of students which was expressed in diaries and determined changes in emotion. Emotional status was extracted from sentiment analysis system based on Plutchik's Wheel of Emotions and expressed through time. They emphasized that the system was very useful and had potential impacts on communication between instructor and student. \\
Zied et al.  \cite{c7} published a study on improving e-learning teaching quality using sentiment analysis. In the study, they used features selection methods which were Mutual Information, Information Gain, Chi-Square and applied to Hidden Markov Model with Support Vector Machine. They asserted that Information Gain obtained the best result for sentiment analysis in term of F1-Score. Besides, Mutual Information performed better than Chi-Square. \\
Noteworthy, in Vietnamese, Hung T.Vo et al. \cite{c8} used machine learning techniques to categorize sentiment and topic. The data used was collected from eight semesters, including all the company's appraisal responses after the internship. In this study, the author evaluated and analyzed the results on three classification models, Naive Bayes, K-Nearest Neighbors and Support Vector Machine, along with the proposed Bag-of-Structure feature. Support Vector Machine model yielded the highest result on the f1-score with 77.3\% in sentiment classification task and 87.3\% in topic classification task.
\\ Despite the great success of sentiment analysis in the field of education, there are some defects in previous studies. We realized that the studies on the Vietnamese language are relatively few, mainly the research in others \cite{c9,c10}. This happens due to the limitation of datasets in Vietnamese, especially in the field of education. In addition, there is no application of deep learning in the field of sentiment analysis for education, although they are widely used and achieves quite good results. Based on previous studies, the goal of this research is to improve the quality of training activities in the institutions through analyzing student feedback. Firstly, we conduct experiments and analyzing the results to choose the most effective model for applying in our application. Second, we attempt to build an application that automatically classifies student feedback. 

\section{DATA}
  The dataset used in this study is UIT-VSFC: Vietnamese Students’ Feedback Corpus for Sentiment Analysis \cite{c23} which contains more than 16.000 feedback collected from 2013 – 2016 at a university. The dataset is labeled by two parts, which are sentiment (positive, negative, neutral) and topic (curriculums, lecturers, facilities, others). The feedback is collected from an automated survey system, at the end of every semesters. There are two parts, including evaluating given criteria on the 5-point scale and giving feedback directly about the shortcomings of the institution.

\begin{table}[htb]
\centering
\caption{Distributions of sentiment labels according to sentence feedback length}
\begin{tabular}{|l|r|r|r|r|}
\hline
                 & \multicolumn{1}{c|}{Positive} & \multicolumn{1}{c|}{Negative} & \multicolumn{1}{c|}{Neutral} & \multicolumn{1}{c|}{Overall} \\ \hline
1-10             & 25.2                          & 16.6                          & 2.9                          & 44.8                         \\ \hline
10-20            & 19.4                          & 17.2                          & 1.0                          & 37.7                         \\ \hline
20-30            & 1.1                           & 4.8                           & 0.1                          & 6.1                          \\ \hline
\textgreater{}30 & 4.1                           & 7.2                           & 0.4                          & 11.7                         \\ \hline
Overall          & 49.8                          & 45.8                          & 4.3                          & 100                          \\ \hline
\end{tabular}
\end{table}
Table 1 describes the distributions of sentiment labels according to sentence feedback length. As we can see, positive feedback account for 49.8\% as the largest proportion, followed by the negative and neutral feedback. This is because of the characteristics of student feedback mainly focus on 2 main sentiment labels which are positive (49.8\%)  and negative (45.8\%), meanwhile neutral label account for small proportion (4.3\%). \\

\begin{table}[htb]
\begin{center}
\caption{Distributions of topic labels according to sentence feedback length}
    \begin{tabular}{ | l | r | r |r | r | r |}
    \hline
      & Lec. & Cur. & Fac. & Oth. & Overall \\ \hline
    1-10 & 33.0 & 6.7 & 1.5 & 3.4 & 44.8 \\ \hline
    10-20 & 27.6 & 7.3 & 1.7 & 1.0 & 37.7 \\ \hline
    20-30 & 7.6 & 2.8 & 0.6 & 0.2 & 11.3 \\ \hline
   $\geq30$ & 3.5 & 1.9 & 0.4 & 0.1 & 6.1 \\ \hline
    Overall & 71.7 & 18.8 & 4.4 & 5.0 & 100 \\
    \hline
    \end{tabular}
\end{center}
\end{table}
Table 2 demonstrates the distributions of sentiment labels according to sentence feedback length. Lecturers feedback account for the largest proportion of 71.7\%, followed by curriculums (18.8\%), others (5.0\%) and facilities (4.4\%). \\
Consider the sentence length, the feedback ranging from 1 to 10 accounts for the largest proportion with 44.8\%. On the other hand, the feedback ranging from 20-30 accounts for the lowest rate of 6.1\%. In general, the corpus is imbalanced in both sentiment part and topic part. This will negatively affect the results of classification models.

\section{METHODOLOGY}
There have been many classification methods used in sentiment analysis which can be applied to student feedback. In this study, we will conduct experiments, analyze the final results on two traditional classifiers (Naive Bayes, Maximum Entropy) and two deep learning classifiers (Long Short-Term Memory, Bi-Directional Long Short-Term Memory) to choose the most effective model for our application.   \\
\subsection{\textit{Traditional Classifiers}}
\begin{itemize}
\item Naive Bayes (NB)
\end{itemize}
Naive Bayes is a simple algorithm, is used to solve problems related to data classification based on statistical methods. With assumptions of independence assuming the dominant role, Naive Bayes concludes faster than logistic regression models \cite{c20}. The main disadvantage of Naive Bayes is that it can not find the interplay between features. 
Its classifier is built on the foundation of mathematical formulas statistical probability of the same name. Suppose we give the data \(d\) and classify \(c\) into the Naive Bayes formula, as follows:
\begin{equation}
{P(c|d)} = \frac{P(d|c)P(c)}{P(d)}
\end{equation}
The problem is that for any given input \(d\), the model will determine which \(c\) class is appropriate. We have the formula for calculating the probability \(c\) of class \(d\):
\begin{equation}
{C_{MAP}} = argmax_{c\in{C}}P(d|c)P(c)
\end{equation}

With \(C_{MAP}\) is the probability of class \(c\) can occur with \(d\). \\
We applied 4 features which are uni-gram, bi-gram, dependency relation (DEP) and POS-tag (POS).

\begin{itemize}
\item Maximum entropy (Maxent)
\end{itemize}
The principle of maximum entropy states that the probability distribution which best represents the current state of knowledge is the one with the largest entropy, in the context of precisely stated prior data [11]. Maxent tries to maximize the entropy by estimating the probabilities of labels based on the features of the sentence. Maxent use features to set conditional distribution. The probability of class \(c\) given the document \(d\) and weight vector \(\lambda\) 
is :
\begin{equation}
P(c|d) = \frac{exp(\sum_{i=1}^{k} \lambda_if_i(c,d))}{\sum_{c'}^{N}exp\sum_{i}^{k}\lambda_if_i(c',d)}
\end{equation}

In above formula, \(k\) is the number of features in the sentence, \(\lambda\) is the weight vector, the higher \(\lambda\)  corresponding to the higher weight. \(\lambda\) is optimized though Conjugate Gradient and Smoothing method or though Quadratic Penalties. A feature function in Maxent model is represented as \({f_{i}(c,d)}\) , where \(c\) is the class and \(d\) is the document. \({f_{i}(c,d)}\) is represented by: \\

\begin{equation}
    {f_{i}(c,d)}=\left\{
                \begin{array}{ll}
                  1, \quad c = c_i \ \textrm{and} \ d = d_i\\
                  0, \quad \textrm{otherwise}
                \end{array}
              \right.
\end{equation}

We applied 4 features which are uni-gram, bi-gram, dependency relation (DEP) and POS-tag (POS).

\subsection{\textit{Deep Learning}}
\begin{itemize}
\item Long Short-Term Memory
\end{itemize}
Long Short-Term Memory(LSTM) \cite{c22} is a variant of Recurrent Neural Network (RNN) which has been widely used for classifications. A node in LSTM consists of a cell, an input gate and each gate has a sigmoid layer and a pointwise multiplication operation.
Firstly, LSTM based on \(h_{t-1}\) and \(x_t\) to eliminate some information in cell state by outputting a number between 0 and 1 with each \(C_{t-1}\).
\begin{equation}
f_t = \sigma(W_fx_t + U_f + b_f) 
\end{equation}

Next the LSTM will calculate the value for the \(i_t\)  and produce a candidate \(\tilde{C_t}\) value.
\begin{equation}
i_t = \sigma(W_ix_t + U_ih_{t-1} + b_i) 
\end{equation}

\begin{equation}
\tilde{C_t} = tanh(W_cx_t +	U_ch_{t-1} + b_c)
\end{equation}

Then the cell state at the current time will be updated through \(C_{t-1}\) and  \(\tilde{C_t}\)
\begin{equation}
C_t = i_t\tilde{C_t} + f_tC_{t-1}
\end{equation}

Finally, \(o_t\) is calculated to determine which part of \(C_t\) will be selected as the output and output will be decided via \(C_t\)  and a \(tanh\) layer (produces a value between -1 and 1).
\begin{equation}
	o_t = \sigma(W_ox_t + U_oh_{t-1} + V_oC_t + b_o) \\
\end{equation}

\begin{equation}
	h_t = o_ttanh(C_t)
\end{equation}

The input of the LSTM used was Word Embedding (Word2Vec model) which was trained on our collected student feedback data. \\

\begin{itemize}
\item Bi-Directional Long Short-Term Memory (Bi-LSTM)
\end{itemize}
Bi-LSTM is an extension of the traditional LSTM network, which has the ability to learn more input information. The LSTM's operational mechanism is based on the main idea of Bi-Directional Recurrent Neural Network \cite{c18}. The idea is to split the state neurons of a regular LSTM in two parts, one is responsible for the positive time direction (forward states) and the other is responsible for the negative time direction (backward states). Outputs from forward states are not connected to inputs of backward states, and vice versa \cite{c18}. In other words, Bi-LSTM has the ability to learn more contextual information extracted from two directions.

\begin{figure}[ht]
  \centering
    \includegraphics[width=0.4\textwidth]{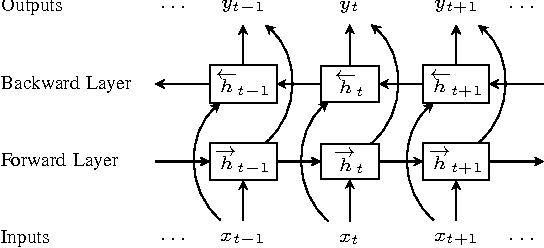}
    \caption{Bi-Directional Long Short-Term Memory structure \cite{c18}}
\end{figure}

As illustrated above, the Bi-LSTM network consists of two hidden states \overleftarrow{h} and \overrightarrow{h} which are calculated as a regular LSTM network. The output at step \(t\) is calculated based on the combination of \overleftarrow{h} and \overrightarrow{h}.\\

The input of the Bi-LSTM used was Word Embedding (Word2Vec model) which was trained on our collected student feedback data. \\

\section{EXPERIMENTAL SETTINGS }
Experiments were conducted to better understand the operation mechanism of the four main algorithms: Naive Bayes, Maximum Entropy, Long Short-Term Memory and Bi-Directional Long Short-Term Memory. As stated above, we used UIT-VSFC on Vietnamese containing more than 16.000 labeled by topic and sentiment. All 4 algorithms were implemented on Java frameworks. Before classification, we conducted data preprocessing for next steps. This process including tokenization, lowercasing, removing number, stop words, and irrelevant contents (emoticons, punctuation, etc,. ) to achieve better results and lower the error rate of data. \\
The Naive Bayes algorithm was implemented on the Datumbox framework \cite{c12}. In particular, this classifier is applied Multinomial Naive Bayes Model with chi-square feature selection. We used Standford Classifier \cite{c13} for Maximum Entropy algorithms. This is is a general purpose classifier - something that takes a set of input data and assigns each of them to one of a set of categories. The Long Short-Term Memory algorithm and Bi-Directional Long Short-Term Memory were implemented on the DeepLearning4j framework \cite{c14}. Deeplearning4j is a library written in the Java programming language and supports the development, training, and deployment of deep learning algorithms. To extract dependency relation and get POS tag of feedback, we used a Vietnamese based dependency parsing tool \cite{c19}. Two deep learning models share the same configuration. We train them with 2 layers, 128 units each and 300-dimentional embeddings. We train for 10 epochs and use Stochatic Gradient Descent with the learning rate of 0.02 for optimization. We also apply the dropout technique and set them to 0.4.\\
To evaluate the efficiency of the classifiers, we used the confusion matrix on the test datasets that was known result in advance. To calculate the performance of the classifiers, we used three common measures: precision (P), recall (R), f1 - Score (F1).   All measurements were calculated using the Micro-Average method based on the confusion matrix \cite{c15}. We divided the datasets by the 80:20 corresponding to training data and testing data. More specific, we used 12800 feedback for training and 3200 feedback for testing.\\

\section{EXPERIMENTAL RESULTS}
\subsection{Sentiment}

\begin{table}[htb]
\centering
\caption{Evaluation of sentiment classification task on VSFC(\%)}
\label{my-label}
\begin{tabular}{ | l | l | r | r | r |}
\hline
Algorithms              & Features                  & P             & R             & F1            \\
\hline
\multirow{4}{*}{NB}     & Uni-gram                     & 86.1          & 84.6          & 85.3          \\
                        & Bi-gram                  & 88.3          & 86.8          & 87.5          \\
                        & Bi-gram + DEP             & 88.2          & 86.8          & 87.5          \\
                        & Bi-gram+DEP+POS           & 84.2          & 86.6          & 85.3          \\
\hline
\multirow{4}{*}{Maxent} & Un-igram                     & 83.3          & 84.9          & 84.1          \\
                        & Bi-gram                   & 87.5          & 87.3          & 87.4          \\
                        & Bi-gram+DEP               & 87.6          & 87.5          & 87.5          \\
                        & Bi-gram+DEP+POS           & 87.4          & 87.3          & 87.3          \\
\hline
LSTM                    & Word2Vec & 88.4          & 87.1          & 87.6          \\
\hline
Bi-LSTM                  & Word2Vec  & \textbf{90.8} & \textbf{93.4} & \textbf{92.0} \\ 
\hline
\end{tabular}
\end{table}

Table 3 compares the results of algorithms with their features on the sentiment task. As can be seen, experimental results show that Bi-LSTM with Word2Vec model is always superior to others in all cases. For example, Bi-LSTM achieves the best on precision, recall, and f1-scores which are 90.8\%, 93.4\%, 92.0\%, respectively, while the precision, recall f1-scores of three other models are lower. On the f1-score, LSTM achieves 87.6\% with Word2Vec. Maxent performs better with the result of 87.5\% when combined with bigram and dependency relation. Finally, NB achieves optimum results of 87.5\% when combined with bigram feature. According to some previous studies on sentiment analysis, it has been shown that modern deep learning models such as LSTM and Bi-LSTM achieved higher results than other traditional models \cite{c16,c17}. 

\subsection{Topic}
\begin{center}
\begin{table}[htb]
\caption{Evaluation of topic classification task on VSFC(\%)}
\centering
\label{my-label}
\begin{tabular}{|l | l | r | r | r |} \hline
Algorithms              & Features                  & P             & R             & F1            \\ \hline
\multirow{4}{*}{NB}     & Uni-gram                     & 79.9          & 84.6          & 81.2          \\
                        & Bi-gram                  & 84.2          & 86.8          & 85.4          \\
                        & Bi-gram + DEP             & 84.4          & 86.8          & 85.5          \\
                        & Bi-gram+DEP+POS           & 82.8          & 86.6          & 84.6          \\ \hline
\multirow{4}{*}{Maxent} & Uni-gram                     & 83.3          & 83.2          & 83.2          \\
                        & Bi-gram                   & 86.3          & 85.8          & 86.1          \\
                        & Bi-gram+DEP               & 86.7          & 86.3          & 86.6          \\
                        & Bi-gram+DEP+POS           & 86.6          & 85.9          & 86.2          \\ \hline
LSTM                    &  Word2Vec & 88.2          & 87.2          & 87.7          \\ \hline
Bi-LSTM                  &  Word2Vec  & \textbf{89.3} & \textbf{90.0} & \textbf{89.6} \\ \hline
\end{tabular}
\end{table}
\end{center}

On topic classification task, as shown in Table 4, Bi-LSTM performs better than three other models in term of precision, recall and f1-Score which are 89.3\%, 90.0\%, and 89.6\%, respectively. On the f1-Score, LSTM achieves the result of 87.7\% with Word2Vec. Maxent performs better than NB with the result of 86.6\% when combines with bigram and dependency relation. Finally, NB achieves optimum results of 85.5\% when combined with bigram and dependency relation. Another observation from Table 3 and Table 4 is that when using dependency relation as a feature for NB and Maxent, it provides a significant improvement in performance, but when combining all 3 features the overall results drop. \\
After conducting the experiments, We conclude that deep learning algorithms perform better than traditional algorithms. This is because deep learning algorithms have the ability to exploit information about the context of words. In addition, the uniformity of the data has a big influence on the performance of the classification model.
\subsection{Application}
In this section, we describe our sentiment analysis and topic detection application. This application would help the university administrators to classify student feedback automatically, to identify students' views on a particular issue. From there the institution uses these idea to improve quality in the future. The application simplifies the analysis of student feedback by analyzing and depicting the results on the charts. \\
First, after a fixed format file is imported, the application automatically parses the file to get the student feedback. The feedback are passed through a pre-processing data phrase to eliminate misspellings and invaluable units, and a pre-trained classifier. The input of this application is a list of feedback sentences, the output is two labels corresponding to each feedback which are the sentiment label (positive, neutral, negative) and topic label (curriculums, lecturers, facilities and others). The final result will be visualized on charts. \\
The application performs the results into two types of charts: pies charts showing the status for the current semester and line charts showing trends over time. From the pie chart, the university administrators can also see the specific results of the student's view in the current semester. For example, the rate of feedback on facilities is low, indicating that facilities have been improved. From a line chart, the university administrators can capture trends in students' views on issues in the school. For example, the positive view of the topic of the lecturer over time is increasing, the manager knows that the institution is improving well and is in line with the needs of the students. \\
This application will help the institution to recognize students' opinions about a problem and identify shortcomings that still exist. From there, the institution can propose appropriate improvement method to improve the quality of training activities in the next semesters. \\
\begin{figure*}[h]
  \centering
    \includegraphics[width=1.0\textwidth]{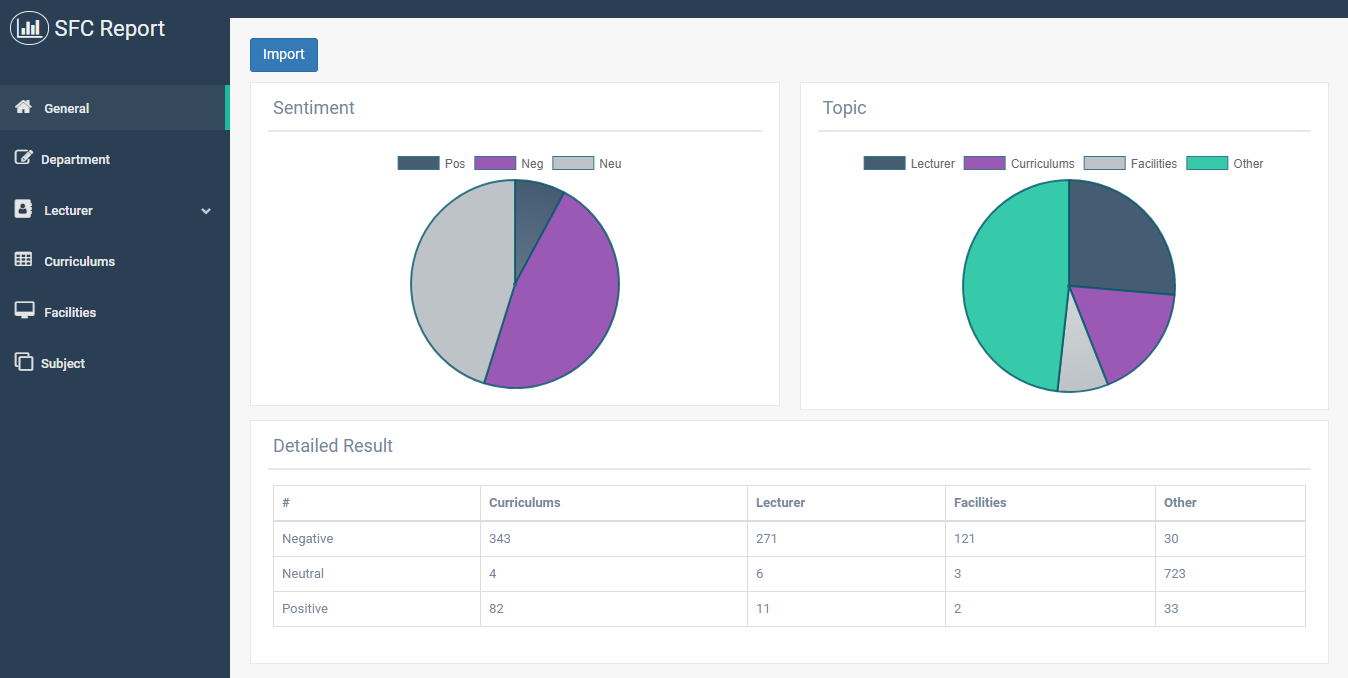}
    \caption{Application home page}
\end{figure*}

\section{CONCLUSION AND FUTURE WORKS}
In this study, we conducted a sentiment analysis of datasets generated from student feedback using supervised learning algorithms. Specifically, our study employs four algorithms including Naive Bayes, Maximum Entropy, Long Short-Term Memory and Bi-Directional Long Short-Term Memory on student feedback datasets. After comparing the results, we conclude that Bi-LSTM outperforms in the two tasks. It achieved 92.0\% in sentiment classification task and 89.6\% in topic classification task in term of F1-score. In addition, we developed an application for analyzing student feedback and providing overview reports for the administrator to identify and recognize student interests in the institution. The system allows the administrator to synthesize the results, perform the statistics and make reports, avoid the status of manual assessment, so that institution can use useful information collected to improve the quality of teaching and learning. \\

For future works, we intend to improve the results of algorithms by adding other features or applying other algorithms. In addition, because of the characteristics of the corpus, we plan to perform unbalanced data-processing techniques to improve the results of small-volume labels. The system needs to be improved in the data preprocessing phase like applying dependency parsing for grammar checking and misspellings detection.

\end{document}